\documentclass{tETA2e}
\usepackage{graphicx}
\usepackage{amsmath}
\usepackage{amssymb}
\usepackage{array}
\usepackage{multirow,tabularx}
\usepackage{epstopdf}
\usepackage{subfigure}
\usepackage{url}

\usepackage[longnamesfirst,sort]{natbib}
\bibpunct[, ]{(}{)}{;}{a}{,}{,}
\renewcommand\bibfont{\fontsize{10}{12}\selectfont}

\theoremstyle{plain}

\theoremstyle{definition}

\theoremstyle{remark}

\newcommand{\bfw}{{\textbf{w}}}
\newcommand{\bfy}{{\textbf{y}}}
\newcommand{\bfg}{{\textbf{g}}}

\newcommand{\bff}{{\textbf{f}}}

\begin{document}



\title{Low-Rank Deep Convolutional Neural Network for Multi-Task Learning}

\author{
\name{Fang Su\textsuperscript{a},
Hai-Yang Shang\textsuperscript{b}
$^{\ast}$\thanks{$^\ast$Corresponding author.},
Jing-Yan Wang\textsuperscript{c}}
\affil{
\textsuperscript{a} Shaanxi University of Science \& Technology, Xi'an, Shaanxi Province, P.R.China, 710021\\
\textsuperscript{b} Lanzhou University of Finance and Economics, Lanzhou ,Gansu Province, P.R.China, 730000\\
\textsuperscript{c} New York University Abu Dhabi, Abu Dhabi, United Arab Emirates}
\received{v5.0 released July 2015}
}

\maketitle

\begin{abstract}
In this paper, we propose a novel multi-task learning method based on the deep convolutional network. The proposed deep network has four convolutional layers, three max-pooling layers, and two parallel fully connected layers. To adjust the deep network to multi-task learning problem, we propose to learn a low-rank deep network so that the relation among different tasks can be explored. We proposed to minimize the number of independent parameter rows of one fully connected layer to explore the relations among different tasks, which is measured by the nuclear norm of the parameter of one fully connected layer, and seek a low-rank parameter matrix. Meanwhile, we also propose to regularize another fully connected layer by sparsity penalty, so that the useful features learned by the lower layers can be selected. The learning problem is solved by an iterative algorithm based on gradient descent and back-propagation algorithms. The proposed algorithm is evaluated over benchmark data sets of multiple face attribute prediction, multi-task natural language processing, and joint economics index predictions. The evaluation results show the advantage of the low-rank deep CNN model over multi-task problems.
\end{abstract}


\section{Introduction}

\subsection{Backgrounds}

{In machine learning applications, multi-task learning has been a popular topic \citep{caruana1997multitask,ruder2017overview,doulamis2016fast,Baniata2018,zhoumultiple,mao2014leave,evgeniou2004regularized,xue2007multi,jacob2009clustered,argyriou2007multi}. }It tries to solve multiple related machine learning problems simultaneously. The motive is that for many situations, multiple tasks are closely related, and the prediction results of different tasks should be consistent. Accordingly, borrowing the prediction of other tasks to help the prediction of a given task is natural. For example, in the face attribute prediction problem, given an image, the prediction of female gender and wearing long hair is usually related \citep{owusu2014svm,yao2018cascade,zhong2016face,cho2008human,wong2010face}. Moreover, in the problem of natural language processing, it is also natural to leverage the problems of part-of-speech (POS) tagging and noun chuck prediction, since a word with a POS of a noun usually appears in a noun chunk \citep{jabbar2018improved,herath1992analysis,collobert2008unified,lyon1997using,shams2016supervised}. Multi-task learning aims to build a joint model for multiple tasks from the same input data.

{In recent years, deep learning has been proven to be the most powerful data representation method \citep{guo2016deep,langkvist2014review,voulodimos2018deep,Chu2018,Chao2018,Sadouk2018,Voulodimos2018,Hu2018,wulearning,zhang2018cross,geng2017novel,zhang2017learning,lecun2015deep,schmidhuber2015deep,ngiam2011multimodal,glorot2011domain}. }Deep learning methods learn a neural network of multiple layers to extract the hierarchical patterns from the original data, and provide high-level and abstractive features for the learning problems. For example, for the face recognition problems, a deep learning model learn simple patterns by the low-level layers, such as lines, edges, circles, and squares. In the median-level layers, parts of faces are learned, such as eyes, noses, mouths, etc. In the high-level layers, patterns of entire faces of different users are obtained. With the deep learning model, we can explore the hidden but effective patterns from the original data directly with multiple layers, even without domain knowledge and hand-coded features. This is a critical advantage compared to traditional shallow learning paradigms models.

{\textbf{Remark}: If shallow learning paradigms is applied in this case, the model structure will not be sufficient to extract complex hierarchical features. The users of these shallow learning models have to code all these complex hierarchical features manually in the feature extraction process, which is difficult and some times impossible.}

{\textbf{Remark}: If other non-neural networks learning models is used, such as spectral clustering, the hidden pattern of input data features cannot be directly explored. For example, spectral clustering treats each data point as a node in a graph, and sperate them by cutting the graph. However, it still needs a powerful data representation method to build the graph, and itself cannot work well with a high-quality graph. Meanwhile, neural network models, especially deep neural network models, have the ability to represent the hidden patterns of input data points and build the high-quality graph accordingly. Thus the non-neural network models and neural network models are complementary.
}
Most recently, deep learning has been found very effective for multi-task learning problems. For example, the following works have discussed the usage of deep learning for multi-task prediction.

\begin{itemize}
\item Zhang et al. \citep{zhang2014facial} formulated a deep learning model constrained by multiple tasks, so that the early stopping can be applied to different tasks to obtain good learning convergence. Furthermore, different tasks regarding face images, including facial landmark detection, head pose estimation, and facial attribute detection are considered together by using a common deep convolutional neural network.

\item Liu et al. \citep{liu2015representation} proposed a deep neural network learning method for multi-task learning problems, especially for learning representations across multiple tasks. The proposed method can combine cross-task data, and also regularize the neural network to make it generalized to new tasks. It can be used for both multiple domain classification problems and information retrieval problems.

\item Collobert et al. \citep{collobert2008unified} proposed a convolutional neural network for multi-task learning problem in natural language processing applications. The targeted multiple tasks include POS tagging, noun chunk prediction, named entity recognition, etc. The proposed network is not only applied to multi-task learning, but also applied to semi-supervised learning, where only a part of the training set is labeled.

\item Seltzer et al. \citep{seltzer2013multi} proposed to learn a deep neural network for multiple tasks which shares the same data representations. This model is used to the applications of acoustic models with a primary task and one or more additional tasks. The tasks include phone labeling, phone context prediction, and state context prediction.
\end{itemize}

However, the relation among different tasks is not explored explicitly. Although the deep neural model can learn effective high-level abstractive features, without explicitly explore the relation of different tasks, different groups of level features may be used to different tasks. Thus the deep features are separated for different tasks, and the relationships among different tasks are ignored during the learning process of the deep network. To solve this problem, we propose a novel deep learning method by regularizing the parameters of the neural network regarding multiple tasks by low-rank.

\subsection{Our Contributions}

The proposed deep neural network is composed of four convolutional layers, three max-pooling layers, and two parallel fully connected layers. The convolutional layers are used to extract useful patterns from the local region of the input data, and the max-pooling layers are used to reduce the size of the intermediate outputs of convolutional layers while keeping the significant responses. The last two fully connected layers are used to map the outputs of convolutional and max-pooling layers to the labels of multiple tasks.

The rows of the transformation matrices of the full connection layers are corresponding to the mapping of different tasks. We assume that the tasks under consideration are closely related, thus the rows of the transformation matrices are not completely independent to each other, thus we seek such a transformation matrix with a minimum number of independent rows. We use the rank of the transformation matrix to measure the number of the independent rows and measure it by the nuclear norm. During the learning process, we propose to minimize the nuclear norm of one fully connected layer's transformation matrix. Meanwhile, we also assume that for a group of related tasks, only all the high-level features generated by the convolutional layers and max-pooling layers are useful. Thus it is necessary to select useful features. To this end, we propose to seek sparse rows for the second fully connected layer. The sparsity of the second transformation matrix is measured by its $\ell_1$ norm, and we also minimize it in the learning process. Of course, we hope the predictions of the two fully connected layers could be low-rank and sparse simultaneously, and also consistent with each other. Thus we propose to minimize the squared $\ell_2$ norm distance between the prediction vectors of the two fully connected layers. Meanwhile, we also reduce the prediction error and the complexity of the filters of the convolutional layers measured by the squared $\ell_2$ norms. The objective function is the linear combination of these terms.

We developed an iterative algorithm to minimize the objective function. In each iteration, the transformation matrices and the filters are updated alternately. The transformation matrices are optimized by the gradient descent algorithm, and the filters are optimized by the back-propagation algorithms.

\subsection{Paper Origination}

The rest parts of this paper are organized as follows. In section \ref{sec:method}, we introduce the proposed method by modeling the problem as a minimization problem and develop an iterative algorithm to solve it. In section \ref{sec:conclusion}, we conclude the paper.

\section{Proposed Method}
\label{sec:method}

\subsection{Problem Modeling}

Suppose we have a set of $n$ data points for the training process, denoted as $\{x_1, \cdots, x_n\}$, where $x_i$ is the $i$-th data point. $x_i$ could be an image (presented as a matrix of pixels) or text (a sequence of embedding vectors of words). The problem of multi-task learning is to predict the label vectors of $m$ tasks. For $x_i$, the label vector is denoted as $\bfy_i = [y_{i1}, \cdots, y_{im}]^\top \in \{1,-1\}^{m}$, where $y_{ij} = 1$ if $x_i$ is a positive sample for the $j$-th task, and $y_{ij} = -1$ otherwise.

\begin{figure}[!htb]
\centering
\includegraphics[width=\textwidth]{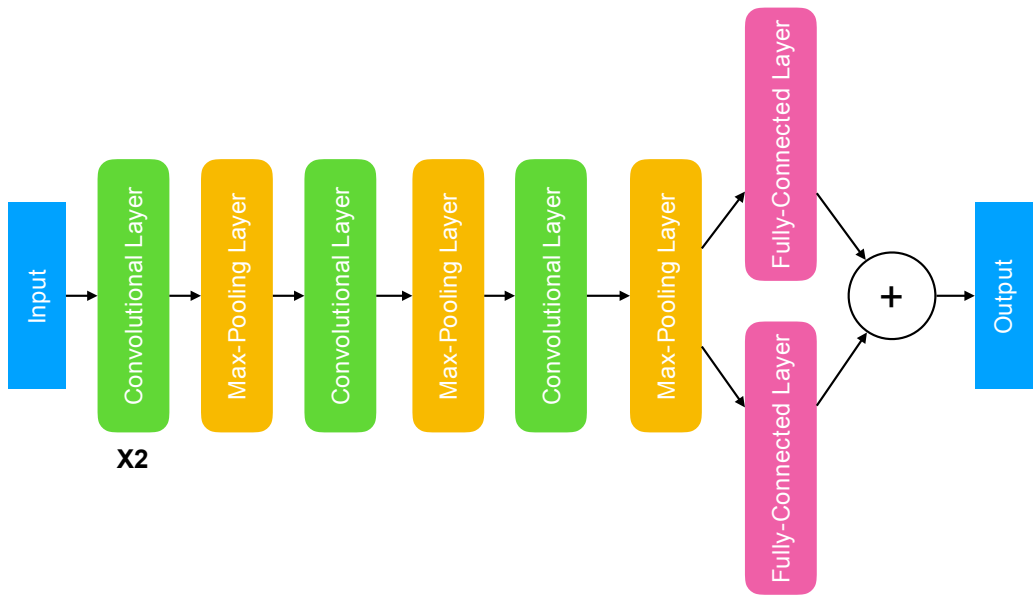}\\
\caption{Sturcture of the proposed deep convolutional network.}
\label{fig:cnn}
\end{figure}

To this end, we build a deep convolutional network to map the input data point to an output label vector. The network is composed of 4 convolutional layers, 3 max-pooling layers, and 2 parallel fully-connected layers. The structure of the deep network is given in Figure \ref{fig:cnn}. Please note that for different types of input data, the convolutional and max-pooling layers are adjustable. For matrix inputs such as images, the layers perform 2-D convolution and 2-D max-pooling, while for sequences such as text, the layers conduct 1-D convolution and 1-D max-pooling.

We denote the intermediate output vector of the first 7 layers as $\phi(x) \in R^p$, where $x$ is the input, and $p$ is the number of pools of the last max-pooling layer. The set of filters in the convolutional layers of $\phi(x)$ are denoted as $\Phi$. The outputs of the two parallel fully connected layers are denoted as

\begin{equation}
\begin{aligned}
\bff_1(x) = W_1 \phi(x) \in R^m, ~and~ \bff_2(x) = W_2 \phi(x) \in R^m,
\end{aligned}
\end{equation}
where $W_1 \in R^{m\times p}$ and $W_2 \in R^{m\times}$ are the transformation matrix of the two layers. The two fully connection layers map the a $p$-dimensional vector $\phi(x)$ to two vectors of $m$ scores for $m$ tasks. Each score measures the degree of the given data point belonging to the positive class. The two fully connected layers are corresponding to the low-rank and sparse prediction results of the network. By fusing their results, we can explore both the low-rank structure of the prediction scores of multi-tasks, and also the sparse structure of the deep features learned from the network. In our model, the first fully connected layer $\bff_1(x)$ is responsible for the low-rank structure, while the second fully connected layer $\bff_1(x)$ is responsible for the sparse structure.

The final outputs of the network are the summation of the outputs of the two fully connected layers,

\begin{equation}
\begin{aligned}
\bfg_1(x) = W_1 \phi(x) + W_2 \phi(x) \in R^m.
\end{aligned}
\end{equation}
To learn the parameters of the deep network of $\bfg_1(x)$, we consider the following four problems.

\begin{itemize}

\item \textbf{Low-Rank Regularization} As we discussed earlier, the tasks are not completely independent from each other, but they are closely related to each other. To explore the relationships between different tasks, we learn a deep and shared representation $\phi(x)$ for the input data $x$. Based on this shared representation, we also request the transformation matrix $W_1$ of one of the last fully connected layer to be of low-rank. The motive is that the $m$ columns of $W_1$ actually map the representation $\phi(x)$ to the $m$ scores of $m$-tasks. The rank of $W_1$ measures the maximum number of linearly independent columns of $W_1$. Thus by minimizing the rank of $W_1$, we can impose the mapping functions of different tasks to be dependent on each other and minimize the number of independent tasks. To measure the rank the matrix $W_1$, $rank(W_1)$, we use the nuclear norm of $W_1$, denoted as $||W_1||_*$. $||W_1||_*$ is calculated as the summation of its singular values,

\begin{equation}
\begin{aligned}
||W_1||_* = \sum_{l} \varrho_l,
\end{aligned}
\end{equation}
where $\varrho_l$ is its $l$-th singular value. We propose to learn $W_1$ by regularizing its rank as follows,

\begin{equation}
\begin{aligned}
\min_{W_1} ||W_1||_*.
\end{aligned}
\end{equation}

\item \textbf{Sparse Regularization} We further regularize the mapping transformation matrix of the second fully connected layer by sparsity. The motive of the sparsity is that the effective deep features for different tasks might be different, and for each task, not all the feature are needed. Although we learn a group of deep features in $\phi(x)$ and share it with all the tasks, for a specific task and its relevant tasks, only a small number of deep features are necessary, and feature selection is a critical step. For the purpose of features selection, we impose the sparsity penalty to the transformation matrix of the second fully connected layer, $W_2$, since it maps the deep features to the prediction scores of $m$ tasks. To measure the sparsity of $W_2$, we use the $\ell_1$ norm of $W_2$, which is the summation of the absolute values of all the elements of the matrix,

\begin{equation}
\begin{aligned}
||W_2||_1 = \sum_{jk} \left| [W_2]_{jk} \right|.
\end{aligned}
\end{equation}
We minimize the $\ell_1$ norm of $W_2$ to learn a sparse $W_2$,

\begin{equation}
\begin{aligned}
\min_{W_2} \frac{1}{2} ||W_2||_1
\end{aligned}
\end{equation}

\item \textbf{Prediction Consistency} The outputs of the two fully connected layers of low-rank and sparsity may give different results. However, we how they can be consistent with each other so that the prediction results can be low-rank and sparse simultaneously. To this end, we impose to minimize the squared $\ell_2$ norm distance between the prediction results of the two layers over all the training data points,

\begin{equation}
\begin{aligned}
\min_{\Phi, W_1, W_2} \frac{1}{2}
\sum_{i=1}^n \left \| W_1 \phi(x_i) - W_2 \phi(x_i) \right \|_2^2.
\end{aligned}
\end{equation}

\item \textbf{Prediction Error Minimization} We also propose to learn an effective multi-task predictor by minimizing the prediction error. To measure the prediction error of a data point, $x$, we calculate the squared $\ell_2$ norm distance between its prediction result $\bfg(x)$ and its true label vector $\bfy$,

\begin{equation}
\begin{aligned}
||\bfy - \bfg(x)||_2^2 = ||\bfy - ( W_1 \phi(x) + W_2 \phi(x) )||_2^2.
\end{aligned}
\end{equation}
We learn the parameters of the deep network by minimizing the errors over all the training data points,

\begin{equation}
\begin{aligned}
\min_{\Phi, W_1, W_2} \frac{1}{2} \sum_{i=1}^n \left \| \bfy_i - (W_1 \phi(x_i) + W_2 \phi(x_i)) \right \|_2^2.
\end{aligned}
\end{equation}

\item \textbf{Complexity Reduction} Finally, we regularize the filters of the convolutional layers, $\Phi$, by the squared $\ell_2$ norms to prevent the network from being over complex,

\begin{equation}
\begin{aligned}
\min_{\Phi} \frac{1}{2}\left \| \Phi \right \|_2^2.
\end{aligned}
\end{equation}

\end{itemize}

The overall optimization problem is the weighted combination of the problems above,

\begin{equation}
\label{equ:object}
\begin{aligned}
\min_{\Phi, W_1, W_2}
& \left\{ g = \frac{1}{2}\left \| \Phi \right \|_2^2
+
\frac{C_1}{2} \sum_{i=1}^n \left \| \bfy_i - (W_1 \phi(x_i) + W_2 \phi(x_i)) \right \|_2^2
\right.\\
&\left. + C_2 ||W_1||_* + \frac{C_3}{2} ||W_2||_1
+ \frac{C_4}{2} \sum_{i=1}^n \left \| W_1 \phi(x_i) - W_2 \phi(x_i) \right \|_2^2
\right \}
\end{aligned}
\end{equation}
where $C_1$, $C_2$, $C_3$ and $C_4$ are the weights of different objective terms, and $g$ is the overall objective function. By optimizing this problem, we can obtain a deep convolutional network with a low-rank and sparse deep features for the problem of multi-task learning.

\subsection{Optimization}

To solve the problem in (\ref{equ:object}), we use the alternate optimization method. The parameters are updated iteratively in an iterative algorithm. When one parameter is updated, others are fixed. In the following sections, we will discuss how to solve them separately.

\subsubsection{Updating $W_1$}

When we update $W_1$, we fix $W_2$ and $\Phi$, remove the terms irrelevant to $W_1$ from (\ref{equ:object}), and obtain the following optimization problem,

\begin{equation}
\label{equ:object}
\begin{aligned}
\min_{W_1}
& \left\{ g_1(W_1) =
\frac{C_1}{2} \sum_{i=1}^n \left \| \bfy_i - (W_1 \phi(x_i) + W_2 \phi(x_i)) \right \|_2^2
\right.\\
&\left. + C_2 ||W_1||_*
+ \frac{C_4}{2} \sum_{i=1}^n \left \| W_1 \phi(x_i) - W_2 \phi(x_i) \right \|_2^2
\right \},
\end{aligned}
\end{equation}
where $g_1$ is the objective function of this problem. To solve this problem, we use the gradient descent algorithm. $W_1$ is descended to the direction of the gradient of $g_1(W_1)$,

\begin{equation}
\label{equ:grident}
\begin{aligned}
W_1 \leftarrow W_1 - \varsigma \nabla g_1(W_1),
\end{aligned}
\end{equation}
where $\nabla g_1(W_1)$ is the gradient function of $g_1(W_1)$, and $\varsigma$ is the descent step size. To calculate the gradient function $\nabla g_1(W_1)$, we first split the objective into two terms,

\begin{equation}
\label{equ:object1}
\begin{aligned}
&g_1(W_1)= g_{11}(W_1) + g_{12}(W_1),~where~\\
&g_{11}(W_1) = \frac{C_1}{2} \sum_{i=1}^n \left \| \bfy_i - (W_1 \phi(x_i) + W_2 \phi(x_i)) \right \|_2^2\\
&+ \frac{C_4}{2} \sum_{i=1}^n \left \| W_1 \phi(x_i) - W_2 \phi(x_i) \right \|_2^2,
\\
&and~g_{12}(W_1) = C_2 ||W_1||_*.
\end{aligned}
\end{equation}
The first term $g_{11}(W_1)$ is a quadratic term while $g_{12}(W_1)$ is a unclear term. Thus the gradient function of $g_1(W_1)$ is the sum of the gradient functions of the two terms,

\begin{equation}
\label{equ:gradient1}
\begin{aligned}
\nabla g_1(W_1)= \nabla g_{11}(W_1) + \nabla g_{12}(W_1).
\end{aligned}
\end{equation}
where $\nabla g_{11}(W_1)$ can be easily obtained as

\begin{equation}
\label{equ:gradient2}
\begin{aligned}
&g_{11}(W_1) = - C_1 \sum_{i=1}^n \left ( \bfy_i - (W_1 \phi(x_i) + W_2 \phi(x_i)) \right )\phi(x_i)^{\top}\\
&+ C_4 \sum_{i=1}^n \left( W_1 \phi(x_i) - W_2 \phi(x_i) \right)\phi(x_i)^{\top}.
\end{aligned}
\end{equation}
To obtain the gradient function of $g_{12}(W_1) = C_2\left \| W_1 \right \|_*$, we first decompose $W_1$ by singular value decomposition (SVD),

\begin{equation}
\label{equ:svd}
\begin{aligned}
W_1 = U \Sigma V,
\end{aligned}
\end{equation}
where $U$ and $V$ are the two orthogonal matrices, $\Sigma$ is a diagonal matrix containing all the singular values. According to the Proposition 1 of \citep{zhen2017multi}, the gradient of $\left \| W_1 \right \|_* =  U \Sigma^{-1} |\Sigma| V$, thus

\begin{equation}
\label{equ:gradient3}
\begin{aligned}
&g_{12}(W_1) = C_2 U \Sigma^{-1} |\Sigma| V.
\end{aligned}
\end{equation}

\subsubsection{Updating $W_2$}

To update $W_2$, we also fix other parameters and remove the irrelevant terms,

\begin{equation}
\label{equ:objectw2}
\begin{aligned}
&g_2(W_2) =
\frac{C_1}{2} \sum_{i=1}^n \left \| \bfy_i - (W_1 \phi(x_i) + W_2 \phi(x_i)) \right \|_2^2
\\
& + \frac{C_3}{2} ||W_2||_1
+ \frac{C_4}{2} \sum_{i=1}^n \left \| W_1 \phi(x_i) - W_2 \phi(x_i) \right \|_2^2\\
&= g_{21}(W_2 )+ g_{22}(W_2),
\end{aligned}
\end{equation}
where

\begin{equation}
\label{equ:objectg21}
\begin{aligned}
g_{21}(W_2 ) = \frac{C_1}{2} \sum_{i=1}^n \left \| \bfy_i - (W_1 \phi(x_i) + W_2 \phi(x_i)) \right \|_2^2
+ \frac{C_4}{2} \sum_{i=1}^n \left \| W_1 \phi(x_i) - W_2 \phi(x_i) \right \|_2^2
\end{aligned}
\end{equation}
is a quadratic term, and

\begin{equation}
\label{equ:objectg21}
\begin{aligned}
g_{22}(W_2) = \frac{C_3}{2} ||W_2||_1
\end{aligned}
\end{equation}
is a $\ell_1$ norm term. We also use the gradient descent algorithm to update $W_2$,

\begin{equation}
\label{equ:gradientw_2}
\begin{aligned}
&W_2 \leftarrow W_2 - \varsigma \nabla g_2(W_2) , ~where~\\
&\nabla g_2(W_2) = \nabla g_{21}(W_2 ) + \nabla g_{21}(W_2 ),~and \\
&\nabla g_{21}(W_2 ) = -C_1 \sum_{i=1}^n \left ( \bfy_i - (W_1 \phi(x_i) + W_2 \phi(x_i)) \right ) \phi(x_i)^{\top}\\
&- C_4 \sum_{i=1}^n \left ( W_1 \phi(x_i) - W_2 \phi(x_i) \right ) \phi(x_i)^{\top}.
\end{aligned}
\end{equation}
To obtain the gradient function of $g_{21}(W_2)$, we rewrite $W_2$ and $\|W_2\|_1$ as follows,

\begin{equation}
\label{equ:w_2}
\begin{aligned}
&
W_2 =
\begin{bmatrix}
\bfw_{21}\\
\vdots\\
\bfw_{2m}
\end{bmatrix}, ~where~
\|W_2\|_1 = \sum_{i=1}^m \| \bfw_{2i} \|_1
= \sum_{i=1}^m \| \bfw_{2i} \|_1,~and \\
&\| \bfw_{2i} \|_1 = \sum_{j=1}^d \left |{\bfw_{2i}}_j\right |
= \sum_{j=1}^d  \frac{{\bfw_{2i}}_j^2}{\left|{\bfw_{2i}}_j\right |} = \bfw_{2i}
diag\left(\left|{\bfw_{2i}}_1\right |, \cdots, \left|{\bfw_{2i}}_d\right |\right)^{-1}
\bfw_{2i}^{\top}
\end{aligned}
\end{equation}
and $\bfw_{2i} = \left [{\bfw_{2i}}_1, \cdots, {\bfw_{2i}}_d\right ]$ is the $i$-th row of $W_2$. The gradient function of $g_{22}(W_2)$ regarding $W_2$, we decompose the problem to the gradients of $g_{22}$ regarding to differen rows of $W_2$, since in the problem the rows are independent to each other,

\begin{equation}
\label{equ:gradientw_2i}
\begin{aligned}
\nabla g_{22}(W_2) =
\begin{bmatrix}
\nabla g_{22}(\bfw_{21})  \\
\vdots\\
\nabla g_{22}(\bfw_{2m}),
\end{bmatrix}
\end{aligned}
\end{equation}
where $\nabla g_{22}(\bfw_{2i})$ is the gradient of $g_{22}$ regarding $\bfw_{2i}$, and according to (\ref{equ:w_2}), we have the sub-gradient of $g_{22}$ as follows,

\begin{equation}
\label{equ:gradientw_2i1}
\begin{aligned}
\nabla g_{22}(\bfw_{2i}) =
C_3 \bfw_{2i}
diag\left(\left|{\bfw_{2i}}_1\right |, \cdots, \left|{\bfw_{2i}}_d\right |\right)^{-1}.
\end{aligned}
\end{equation}

\subsubsection{Updating $\Phi$}

To optimize the filters of the deep network, we fix both $W_1$ and $W_2$ and use the backpropagation algorithm based on the chain rule. The corresponding problem is given as follows,

\begin{equation}
\label{equ:object_phi}
\begin{aligned}
\min_{\Phi}
& \left\{ g_3(\Phi) = \frac{1}{2}\left \| \Phi \right \|_2^2
+  \sum_{i=1}^n \left(
\frac{C_1}{2} \left \| \bfy_i - (W_1 \phi(x_i) + W_2 \phi(x_i)) \right \|_2^2
\right.\right.\\
&\left. \left.
+ \frac{C_4}{2}\left \| W_1 \phi(x_i) - W_2 \phi(x_i) \right \|_2^2
\right )
\right \}\\
&= \frac{1}{2}\left \| \Phi \right \|_2^2 + \sum_{i=1}^n g_{3i}(\Phi),~where\\
&g_{3i}(\Phi) =
\frac{C_1}{2} \left \| \bfy_i - (W_1 \phi(x_i) + W_2 \phi(x_i)) \right \|_2^2\\
&+ \frac{C_4}{2}\left \| W_1 \phi(x_i) - W_2 \phi(x_i) \right \|_2^2
\end{aligned}
\end{equation}
is a data point-wise term. Back propagation is based on gradient descent algorithm,

\begin{equation}
\label{equ:phi}
\begin{aligned}
\Phi \leftarrow \Phi - \varsigma \nabla g_3(\Phi),
\end{aligned}
\end{equation}
and according to the chain rule,

\begin{equation}
\label{equ:chain}
\begin{aligned}
&\nabla g_3(\Phi) =  \Phi + \sum_{i=1}^n \nabla g_{3i}(\Phi),~where \\
&\nabla g_{3i}(\Phi) = \nabla g_{3i}(\phi(x_i)) \nabla_\Phi \phi(x_i),~and\\
&\nabla g_{3i}(\phi(x_i)) = -
C_1 (W_1+W_2)^{\top} \left ( \bfy_i - (W_1 \phi(x_i) + W_2 \phi(x_i)) \right )\\
&+ C_4 (W_1-W_2)^{\top} \left ( W_1 \phi(x_i) - W_2 \phi(x_i) \right ).
\end{aligned}
\end{equation}

%
%

\section{Experiments}

In this section, we test the proposed method over several multi-task learning problems and compare it to the state-of-the-art deep learning methods for the multi-task learning problem.

\subsection{Experiment Setting}

We test the proposed method over the following benchmark data sets.

\begin{itemize}
\item \textbf{Large-scale CelebFaces Attributes (CelebA) Dataset} The first dataset we used is a face image data set, named CelebA Dataset \citep{liu2015faceattributes}. This data set has 202,599 images and each image has 40 binary attributes, such as wearing eyeglasses, wearing hats, having a pointy nose, smiling, etc. The prediction of each attribute is treated as a task, thus this is 40-task multi-task learning problem. The input data is image pixels. The downloading URL for this data set is at \url{http://mmlab.ie.cuhk.edu.hk/projects/CelebA.html}.

\item \textbf{Annotated Corpus for Named Entity Recognition} The second data set we used is a data set for named entity recognition. It contains 47,959 sentences, which contains 1,048,576 words. Each word is tagged by a named entity type, such as Geographical Entity, Organization, Person, etc, or a non-named entity. Moreover, each work is also tagged by a part-of-speech (POS) type, such as noun, pronoun, adjective, determiner, verb, adverb, etc. Meanwhile, we also have the labels of noun chunk. We have three tasks for each work, named entity recognition (NER), POS labeling, and noun clunking. For each word, we use a window of size 7 to extract the context, and the embedding vectors of the words in the window are used as the input. This dataset can be downloaded from \url{https://www.kaggle.com/abhinavwalia95/entity-annotated-corpus}.

\item \textbf{Economics} The third data set we used is a data set for tasks of property price trend and stock price trend prediction. The input data is the wave of historical data of property prices and stock prices, and each data point is the data of three months of both prices of properties and stocks, and the label of each data point is the trend of stock price and property price. We collect the data of last 20 years of USA and China, and generate a total number of 480 data points.
\end{itemize}

In the experiments, we split an entire data set to a training set and a test set of the equal sizes. The training set is used to learn the parameters of the deep network, and then we use the test set to evaluate the performance of the proposed learning method. To measure the performance, we use the average accuracy for different tasks.

\subsection{Experiment Results}

\subsubsection{Comparison of prediction accuracy of different methods}

\begin{figure}
  \centering
  \includegraphics[width=0.8\textwidth]{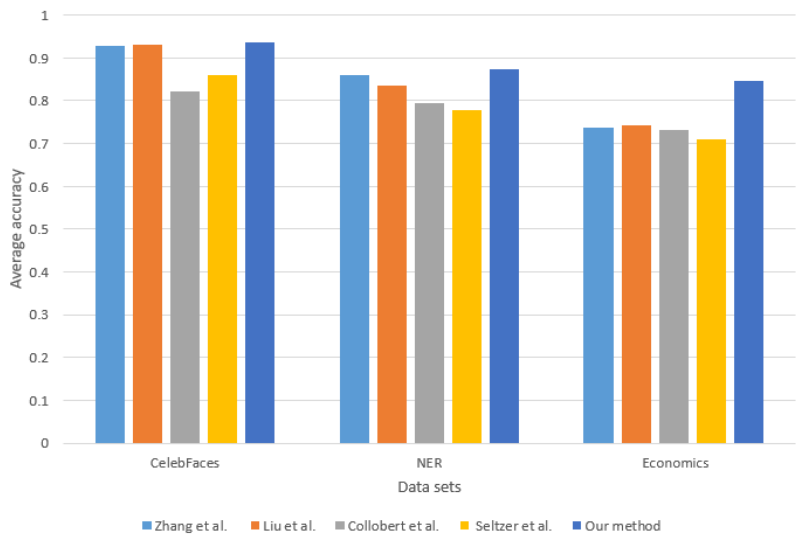}\\
  \caption{Prediction performance of compared methods over benchmark data sets.}
  \label{fig:prediction_acc}
\end{figure}

We compare the proposed method against several deep learning-based multi-task methods, including the methods proposed by Zhang et al. \citep{zhang2014facial}, Liu et al. \citep{liu2015representation}, Collobert et al. \citep{collobert2008unified}, and Seltzer et al. \citep{seltzer2013multi}. The results are reported in Fig. \ref{fig:prediction_acc}. According to the results, the proposed methods always achieves the best prediction performances, over three multi-task learning tasks, especially in the NER and Economics. For the Economics benchmark dataset, our method is the only method which obtains an average prediction accuracy higher than 0.80, while the other methods only obtain accuracies lower than 0.75. This is not surprising since our method has the ability to explore the inner relation between different tasks by the low-rank regularization of the weights of the CNN model for different tasks. In the Economics benchmark data set, the number of training examples is small, thus it is even more necessary to borrow the data representation of different tasks. For the CelebFaces data set, the improvement of the proposed method over the other methods are slight. Moreover, we also observe that the methods Zhang et al. \citep{zhang2014facial} and Liu et al. \citep{liu2015representation} outperforms the methods of Collobert et al. \citep{collobert2008unified}, and Seltzer et al. \citep{seltzer2013multi} in most cases.

\subsubsection{Comparison of running time of different methods}

\begin{figure}
  \centering
  \includegraphics[width=0.8\textwidth]{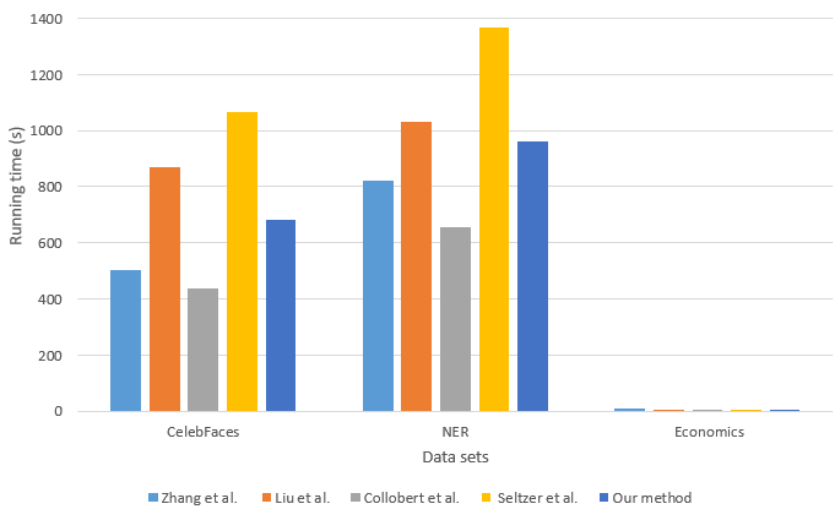}\\
  \caption{Running time of compared methods over benchmark data sets.}
  \label{fig:running_time}
\end{figure}

We also report the running time of the training processes of the compared methods in Fig. \ref{fig:running_time}. According to the results reported in the figure, the training process of Seltzer et al. \citep{seltzer2013multi}'s method is the longest, and the most efficient method is Collobert et al. \citep{collobert2008unified}'s algorithm. Our method's running time of the training process is longer than Zhang et al. \citep{zhang2014facial} and Collobert et al. \citep{collobert2008unified}'s methods, but still acceptable for the data sets of CelebFaces and NER. While for the training process over the Economics benchmark data set, the running time is very short compared to the other two data sets, since its size is relatively small.

\subsubsection{Influence of tradeoff parameters}

\begin{figure}
  \centering
  \includegraphics[width=0.4\textwidth]{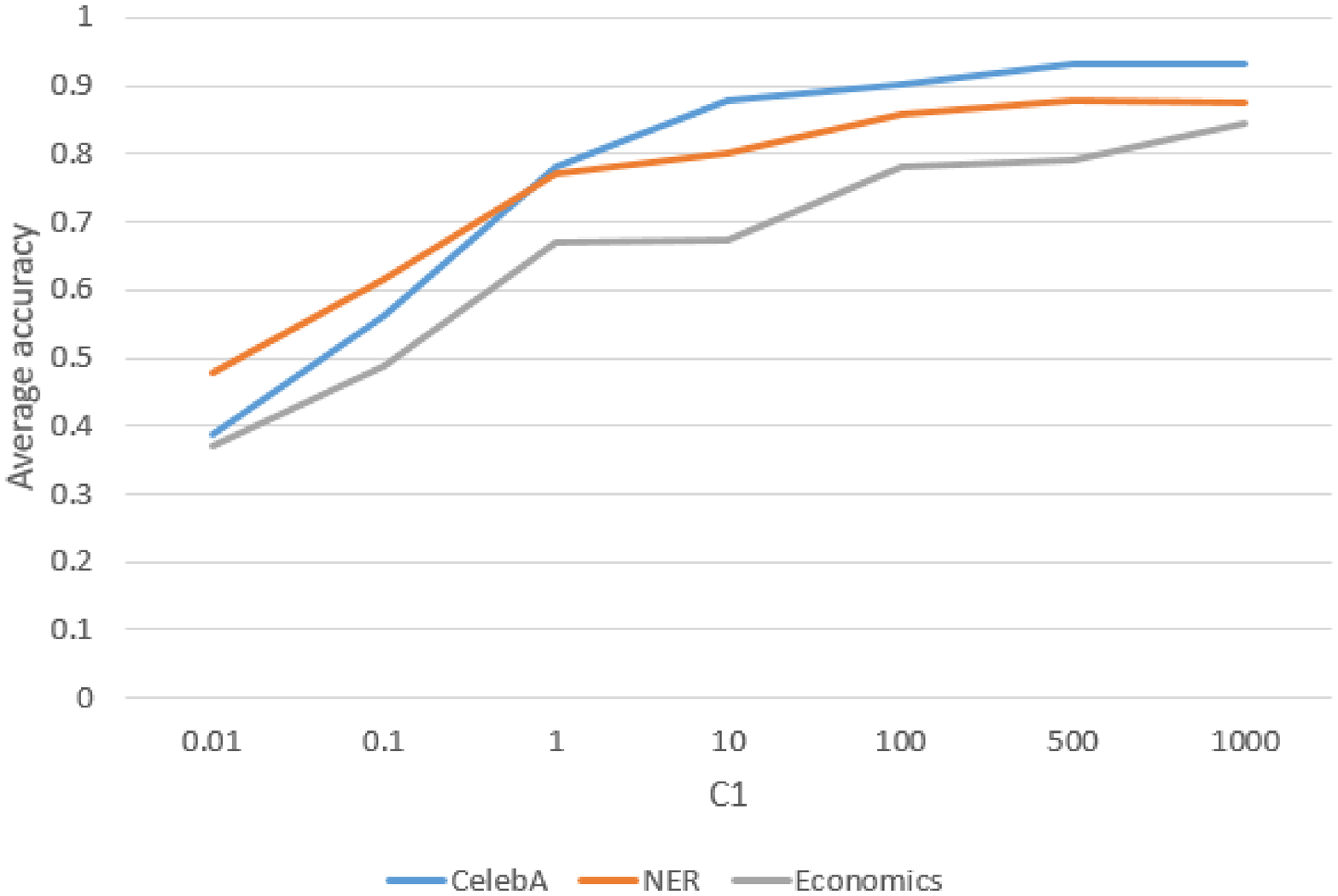}
  \includegraphics[width=0.4\textwidth]{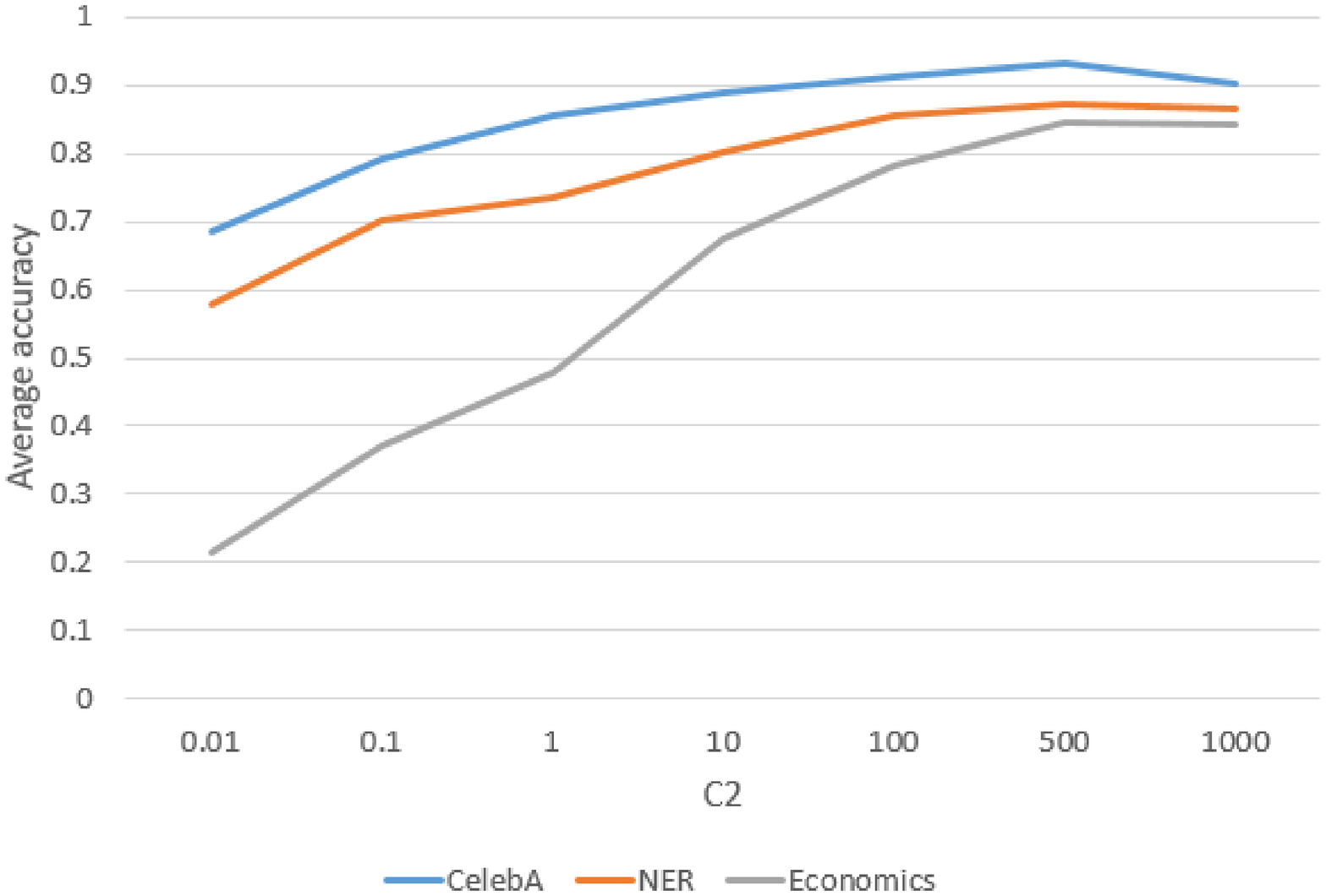}
  \includegraphics[width=0.4\textwidth]{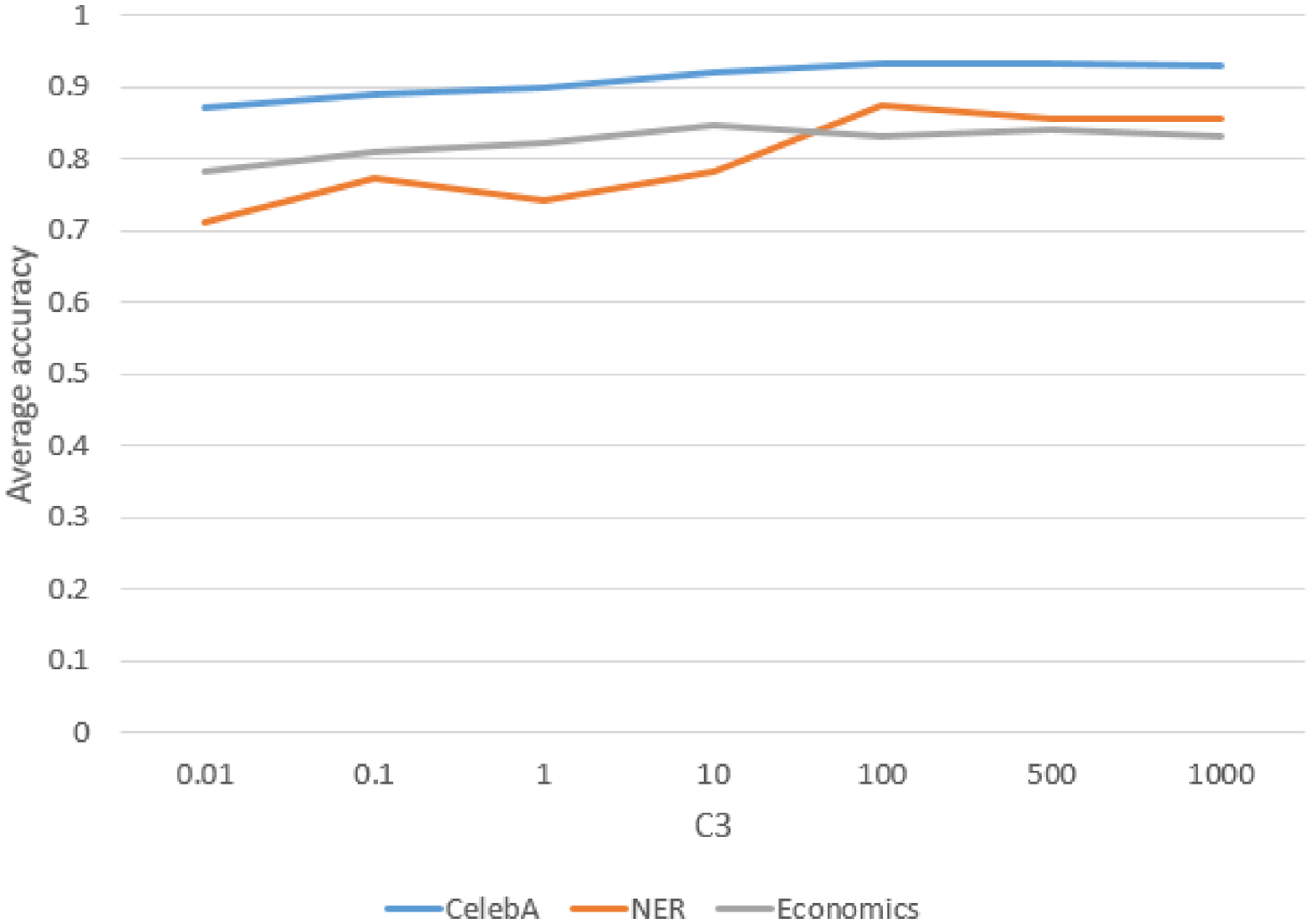}
  \includegraphics[width=0.4\textwidth]{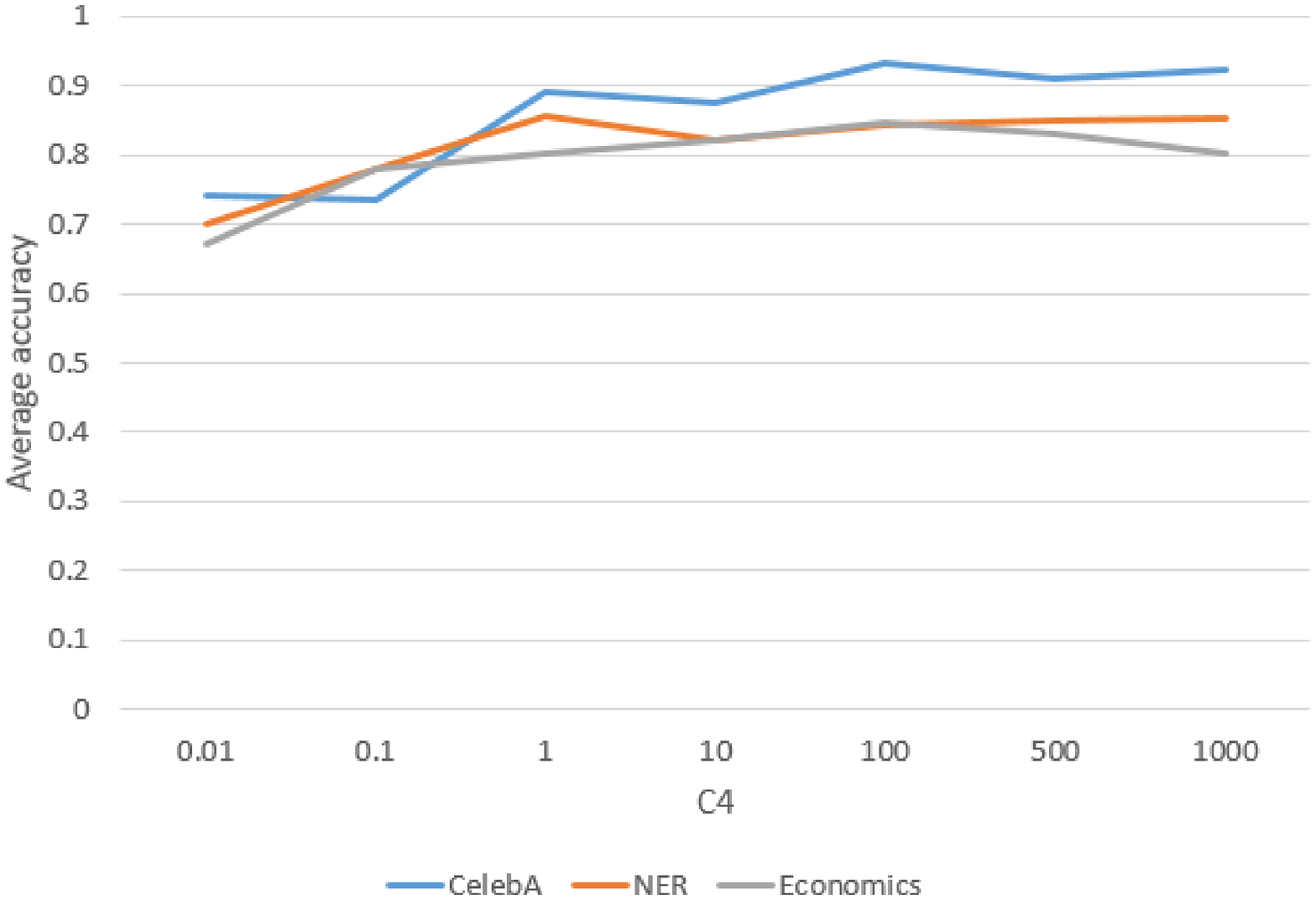}
  \caption{Influences of tradeoff parameters over benchmark data set.}
  \label{fig:tradeoff}
\end{figure}

In our method, there are four important of tradeoff parameters, which controls the weights of the terms of classification errors, the rank of the weight matrix, and the $\ell_1$ norm sparsity of the weight matrix, and the consistency of predictions of the sparse model and low-rank model. The four tradeoff parameters are $C_1$, $C_2$, $C_3$ and $C_4$. We study the influences of the changes of their values to the prediction accuracy and report the results of our method with varying values of these parameters in Fig. \ref{fig:tradeoff}. We have the following observations as follows.

\begin{itemize}
\item According to the results in Fig. \ref{fig:tradeoff}, when the values of $C_1$ increase from 0.01 to 100, the prediction accuracy keeps growing. This is due to the fact that this parameter is the weight of the classification error term, and when its value is increasing the classification error over the training set plays a more and more important role in the learning process, thus it boosts the classification performance accordingly. But when its value is larger than 100, the performance improvement is not significant anymore.

\item When the values of $C_2$ increases, the performance of the proposed keeps improving. This is due to the importance of the low-rank regularization of the proposed method. $C_2$ controls the weight of the low-rank regularization term, and it is the key to explore the relationships among different tasks of multi-task problem. This is even more obvious for the Economics data set, where the data size is small, and cross-task information plays a more important role.

\item The proposed algorithm seems stable to the changes in the values of $C_3$, which is the weight of the sparsity term of the objective. This term plays the role of feature selection over the convolutional representation of the input data. The stability over the changes of $C_3$ implies that the convolutional features extracted by our model already give good performances, thus the feature selection does not significantly improve the performances.

\item For the parameter $C_4$, the average accuracy improves slightly when its value increases until it reaches 100, then the performances seem to decrease slightly. This suggests that the consistency between sparsity and low-rank somehow improves the performance, but it does not always help. For forcing the consistency with a large weight for the consistency term, the performance will not be improved.
\end{itemize}

\section{Conclusion}
\label{sec:conclusion}

In this paper, we proposed a novel deep learning method for the multi-task learning problem. The proposed deep network has convolutional, max-pooling, and fully connected layers. The parameters of the network are regularized by low-rank to explore the relationships among different tasks. Meanwhile, it also has the function of deep feature selection by imposing sparsity regularization. The learning of the parameters are modeled as a joint minimization problem and solved by an iterative algorithm. The experiments over the benchmark data sets show its advantage over the state-of-the-art deep learning-based multi-task models. In the future, we will apply the proposed method to other fields, such as global forest product prediction \citep{liu2018visualized,liu2019study,jiang2019impact}, biomaterials science \citep{zhang2018instructive,zhang2018synthetic,zhang2018peptide,zhang2018vaccine,zhang2018immunomodulatory}, bio-informatics, \citep{gui2016survey,gui2016survey}, wireless networks \citep{liu2014integrated,liu2013local,yang2016efficient,yang2015min}, information communication \citep{wang2016designing,wang2014informing}, etc.


\end{document}